\documentclass[letterpaper, 10 pt, conference]{ieeeconf}  % Comment this line out if you need a4paper

\IEEEoverridecommandlockouts                              % This command is only needed if 
\usepackage{graphicx}
\usepackage[hidelinks]{hyperref}

\usepackage{color,soul}
\usepackage[usenames,dvipsnames]{xcolor}
\usepackage{caption}
\usepackage{subcaption}
\usepackage{multirow}
\usepackage{array}
\usepackage{booktabs}
\usepackage{graphicx}
\usepackage{caption}
\usepackage{tabularx}
\usepackage{float}
\usepackage[normalem]{ulem}
\usepackage{makecell}
\usepackage{placeins}
\usepackage{fancyhdr}

\newcommand{\ieeecopyright}{%
  \footnotesize
  \centering
  \copyright~2026 IEEE.  Personal use of this material is permitted.  Permission from IEEE must be obtained for all other uses, in any current or future media, including reprinting/republishing this material for advertising or promotional purposes, creating new collective works, for resale or redistribution to servers or lists, or reuse of any copyrighted component of this work in other works.
}

\fancypagestyle{ieeefirstpage}{%
  \fancyhf{}% clear all header and footer fields
  \fancyfoot[C]{\ieeecopyright}% centered footer
}

\title{\LARGE \bf
\SYS{}: GPU-Accelerated Local Mapping for Visual SLAM}

\author{Parsa Hosseininejad$^{1}$, Kimia Khabiri$^{1}$, Shishir Gopinath$^{1}$, 
\\ Soudabeh Mohammadhashemi$^{1}$, Karthik Dantu$^{2}$, Steven Y. Ko$^{1}$% <-this % stops a space
\thanks{$^{1}$Simon Fraser University. 
\textcolor{black}\texttt{\{
\href{mailto:sph6@sfu.ca}{\textcolor{black}{sph6}}, 
\href{mailto:kka156@sfu.ca}{\textcolor{black}{kka156}}, \href{mailto:sgopinat@sfu.ca}{\textcolor{black}{sgopinat}}, \href{mailto:sma406@sfu.ca}{\textcolor{black}{sma406}}, \href{mailto:steveyko@sfu.ca}{\textcolor{black}{steveyko}}\}@sfu.ca}}
\thanks{$^{2}$University at Buffalo. 
\texttt{\href{mailto:kdantu@buffalo.edu}{\textcolor{black}{kdantu@buffalo.edu}}}}
}

\newcommand{\SYS}[1]{\textcolor{black} {TurboMap}}
\newcommand{\TUMVI}[1]{\textcolor{black} {TUM-VI}}
\newcommand{\EUROC}[1]{\textcolor{black} {EuRoC}}

\begin{document}

\maketitle
\thispagestyle{ieeefirstpage}
\pagestyle{empty}

\begin{abstract}

In real-time Visual SLAM systems, local mapping must operate under strict latency constraints, as delays degrade map quality and increase the risk of tracking failure. GPU parallelization offers a promising way to reduce latency. However, parallelizing local mapping is challenging due to synchronized shared-state updates and the overhead of transferring large map data structures to the GPU. This paper presents \SYS{}, a GPU-parallelized and CPU-optimized local mapping backend that holistically addresses these challenges. We restructure Map Point Creation to enable parallel Keypoint Correspondence Search on the GPU, redesign and parallelize Map Point Fusion, optimize Redundant Keyframe Culling on the CPU, and integrate a fast GPU-based Local Bundle Adjustment solver. To minimize data transfer and synchronization costs, we introduce persistent GPU-resident keyframe storage. Experiments on the \EUROC{} and \TUMVI{} datasets show average local mapping speedups of $1.3\times$ and $1.6\times$, respectively, while preserving accuracy.

\end{abstract}

\section{Introduction} \label{sec:introduction}

Simultaneous Localization and Mapping (SLAM) systems enable robots to estimate their position and build a map of the environment by processing data from various sensors, including cameras, LiDAR, and inertial measurement units (IMUs). Visual SLAM systems, in particular, rely on visual sensors, typically monocular, stereo, or RGB-D cameras, to reconstruct the environment and localize the robot within it. 

A typical visual SLAM pipeline consists of three main components: tracking, local mapping, and loop closing~\cite{ABASPURKAZEROUNI2022117734}. Tracking estimates the camera pose by extracting features from the current frame and matching them to existing map points, local mapping refines and expands the map to improve accuracy of the system, and loop closing detects revisited locations to correct accumulated drift. Well-known examples of this class of SLAM systems include ORB-SLAM~\cite{Mur-Artal2015, Mur-Artal2017, Campos2021}, VINS-Mono~\cite{VINS-Mono}, and Kimera~\cite{Kimera}.

In real-time visual SLAM, each component must operate within strict time constraints, since delays in any stage degrade overall system performance, with local mapping being particularly critical. 
% \mh{i guess i didnt understand why is this requirement more important for globalmap, its explained later ?}
This requirement is even more pronounced in systems that maintain a global map~\cite{Campos2021}~\cite{Kimera}~\cite{murai2024_mast3rslam}. Such systems generally achieve higher accuracy due to consistent map optimization and loop closure, but suffer from degraded performance in highly concurrent environments due to inefficient concurrent access to the shared global map~\cite{Semenova2024}. As a result, local mapping performance becomes especially important, as falling behind or skipping local mapping leads to less accurate maps and increases the risk of tracking loss.

To mitigate these issues, achieving low latency in local mapping is essential. One promising approach is to parallelize the most computationally intensive tasks, for which GPUs are well suited. However, enabling parallel computation in local mapping is challenging due to the algorithmic structure of visual SLAM systems. 
First, local mapping frequently reads and modifies shared map structures while other modules are running concurrently and accessing the same data. To ensure consistency, these operations use synchronization (e.g., using locks), forcing many computations to wait for exclusive access and thereby
% \mh{serializing gives me a wrong impression, reminds of serialization and deserialization, are you trying to say "forcing sequential execution" or "sequentializing the execution"}
serializing the execution. Consequently, without restructuring the underlying computations, the design of visual SLAM systems inherently limits the effectiveness of parallel execution in local mapping. Second, local mapping operates on large keyframe and map point data structures. Transferring these data to the GPU at each iteration can introduce substantial data movement overhead, which may offset the computational gains of parallel execution if not carefully managed. These challenges explain why prior acceleration efforts \cite{Gopinath2023}\cite{MegBA2022}\cite{CeresCUDA} have mainly focused on components that are naturally data-parallel and do not involve shared-state updates or synchronization. To the best of our knowledge, no previous work addresses the local mapping module holistically while explicitly tackling both synchronization-induced serialization and data transfer overhead.

We propose \SYS{}, a redesigned and accelerated local mapping module that enables efficient parallel execution by addressing two fundamental challenges: synchronization-induced serialization caused by shared-state updates and the high overhead of transferring large data structures to the GPU. Our contributions are threefold: (1) we analyze data dependencies and runtime latency within the local mapping pipeline to identify the components that dominate execution time and evaluate whether their computations can be parallelized; (2) we restructure selected computations to remove synchronization and sequential dependencies, enabling parallel execution on the GPU, while refining CPU-side procedures when parallelization does not provide measurable benefits; and (3) for the components executed on the GPU, we introduce a persistent keyframe storage system that keeps required keyframe data resident on the GPU throughout the local mapping process and stores only immutable data, reducing data transfer and synchronization overhead to improve overall execution efficiency.

We select ORB-SLAM3 as our representative visual SLAM system and implement \SYS{} on top of it to concretely demonstrate the benefits of our approach.

We compare the local mapping performance of \SYS{} with the vanilla ORB-SLAM3 using widely used datasets \EUROC{}~\cite{euroc} and \TUMVI{}~\cite{tumvi}. Our experiments show that \SYS{} achieves an average speedup of $1.3\times$ on \EUROC{} and $1.6\times$ on \TUMVI{} in the local mapping module compared to ORB-SLAM3, across both desktop and embedded platforms, while maintaining similar trajectory accuracy. Although our implementation is based on ORB-SLAM3, 
% \mh{"we propose techniques that generalize"? - more assertive}
we expect the proposed techniques to generalize to
other visual SLAM systems that maintain a global map and implement the components we accelerate in \SYS{}. 
% We will open-source \SYS{} upon acceptance.
Our code is available at \url{https://github.com/sfu-rsl/TurboMap}. 

\section{Related Work} \label{sec:related-work}

In this section, we discuss related work on GPU acceleration for tracking, local mapping, and loop closing modules in the context of ORB-SLAM based systems~\cite{Mur-Artal2015, Mur-Artal2017, Campos2021}.

Several studies focus on optimizing the tracking module in SLAM systems. Aldegheri et al.~\cite{Aldegheri2019} analyze the data flow of ORB-SLAM2 and exploit it to introduce additional levels of parallelism and accelerate feature extraction. Kumar et al.~\cite{KumarParkBehera2024} implement bounded rectification for FAST and pyramidal culling and aggregation and introduce a middle-end component for tracking. Khabiri et al.~\cite{Khabiri2025} use GPU to accelerate key bottlenecks in the ORB-SLAM3 tracking module, including feature extraction, stereo feature matching, and local map tracking, achieving up to $2.8\times$ speed up in the tracking process on both desktop and embedded platforms.

Other works focus on accelerating least squares subproblems in various modules. Gopinath et al.~\cite{Gopinath2023} parallelize matrix calculations to accelerate local bundle adjustment in local mapping, while Kumar et al.~\cite{Kumar2024} offload Jacobian computations for pose graph optimization to the GPU. Ceres Solver~\cite{CeresCUDA} introduces optional CUDA-based linear solvers to speed up the optimization steps in bundle adjustment, and MegBA~\cite{MegBA2022} provides a fully GPU-native, distributed BA framework capable of handling large-scale optimizations.

In contrast to previous work on local mapping that primarily focuses on bundle adjustment \cite{Gopinath2023}\cite{MegBA2022}\cite{CeresCUDA}, we target the remaining components that dominate runtime and are constrained by synchronization and shared-state updates. Specifically, we restructure Map Point Creation and parallelize the Keypoint Correspondence Search stage, and also redesign the Map Point Fusion stage to enable parallel execution. We also optimize Redundant Keyframe Culling on the CPU. To support these parallelized GPU stages, we also design a dedicated keyframe storage system that efficiently manages the data required by these operations. We will introduce these components in \autoref{sec:design}.

\section{Background} \label{sec:background}

\begin{figure}[t]
    \centering
    \includegraphics[width=1\linewidth]{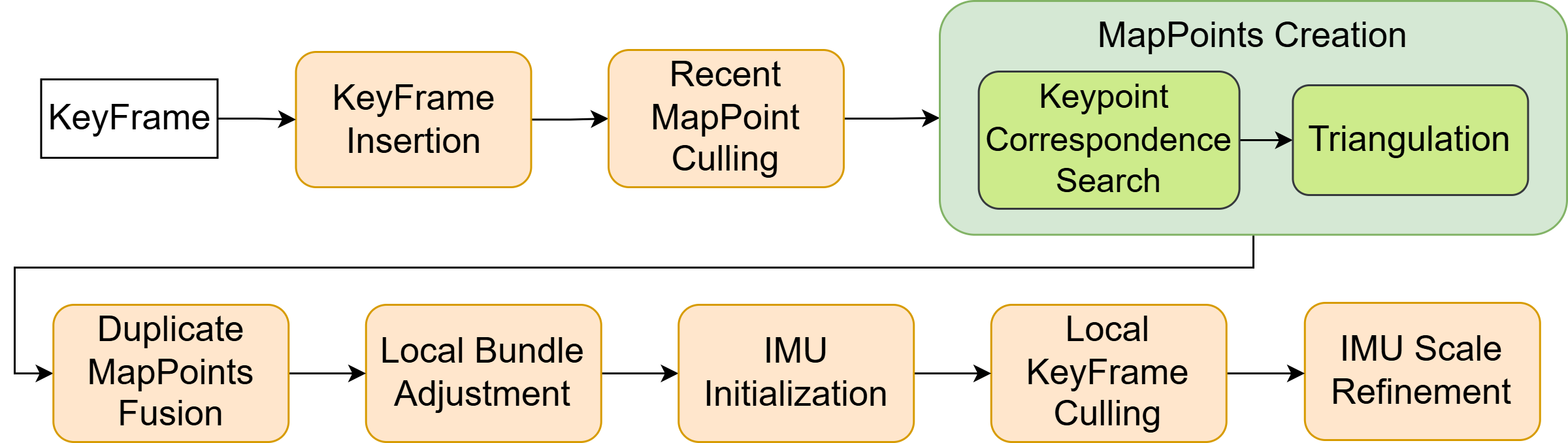}
    \caption{Local mapping workflow in ORB-SLAM3.}
    \label{fig:local_mapping_stages}
    \vspace{-10pt}
\end{figure}

The local mapping module in a visual SLAM system is responsible for processing keyframes as well as building, maintaining, and optimizing the map. The computations performed in this component are critical for ensuring both the efficiency and accuracy of the system. \autoref{fig:local_mapping_stages} illustrates the different stages of local mapping, using ORB-SLAM3 as an example visual SLAM system. In the rest of the section, we explain the components related to our work.

% \mh { you cover 4 sections of the local mapping however its not clear to me why you pick these 4, maybe one sentence saying we describe ... because .. ?}
\subsection{Map-point Creation}
\label{bg:mp_creation}
Map Point Creation is responsible for estimating the 3D coordinates of 2D keypoints and integrating the resulting 3D points into the map. The system begins by performing a keypoint correspondence search across neighboring keyframes to identify matching keypoints. Once the correspondences are found, triangulation is used to determine their 3D positions. After triangulation, a set of geometric and photometric constraints is evaluated, and new map points are inserted into the map only if these constraints are satisfied. The constraints include sufficient parallax between views, positive depth (i.e. points must lie in front of the camera), low re-projection error, and consistency in scale.

\subsection{Map-point Fusion}
\label{bg:mp_fuse}
Map Point Fusion aims to eliminate duplicate map points to maintain a cleaner and more accurate representation of the environment. The process begins by retrieving the neighboring keyframes of the current keyframe, including some second-order neighbors. For each map point in the current keyframe, the system searches for the map point most similar to the map point in each retrieved keyframe. If a pair of map points is determined to be sufficiently similar based on geometric and descriptor-based criteria, they are merged, with one of them being discarded. This fusion process helps produce a sparser and more consistent map.

\subsection{Local Bundle Adjustment}
\label{bg:LBA}
Local bundle adjustment (LBA) refines the poses of local keyframes and the positions of the 3D map points observed by them. This process is typically formulated as a nonlinear least-squares problem that minimizes reprojection residuals (and inertial residuals in visual–inertial systems). The optimization is solved iteratively, often using Gauss–Newton or Levenberg–Marquardt methods. To improve efficiency, the large number of point variables is handled via the Schur complement, which eliminates them from the system of equations so that only the smaller set of pose variables is solved directly. The resulting updates to poses and map points reduce local drift and improve trajectory consistency.

\subsection{Local Keyframe Culling}
\label{bg:kf_culling}
Keyframe culling identifies and removes redundant local keyframes to maintain a compact and efficient map. SLAM systems typically handle this process differently. For example, in ORB-SLAM3, a keyframe is considered redundant if at least 90\% of the map points it observes are also seen in three or more other keyframes at the same or finer scale. Removing such non-informative keyframes reduces memory usage and computational overhead, leading to a more efficient system. This process also benefits local bundle adjustment, whose computational complexity increases with the number of keyframes involved.

\begin{figure}[t]
    \centering
    \includegraphics[width=0.7\linewidth]{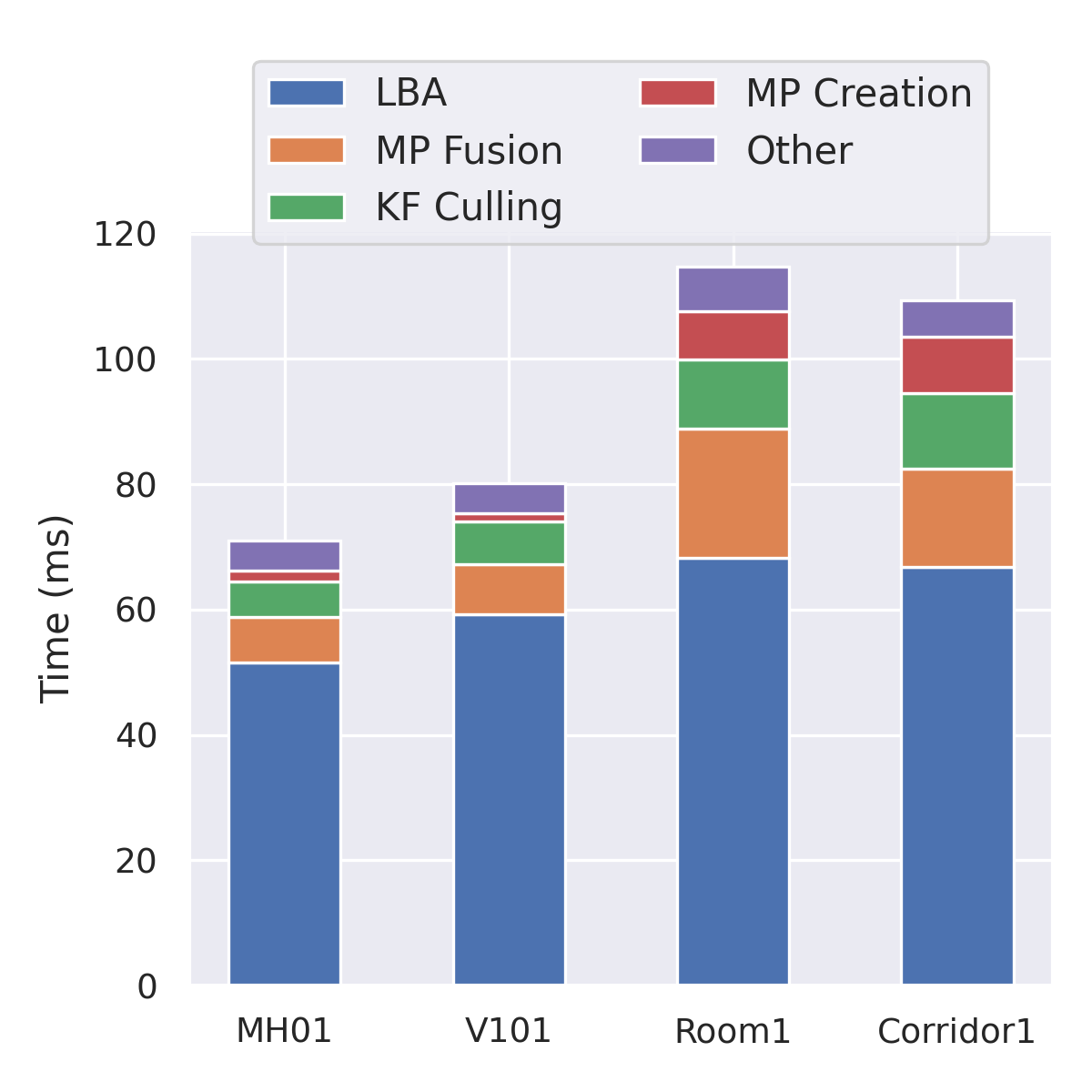}
    \caption{Average time spent in different sections of local mapping in ORB-SLAM3.}
    \label{fig:local_mapping_time_breakdown}
    \vspace{-15pt}
\end{figure}

\begin{figure*}[!t]
    \centering
    \includegraphics[width=0.85\linewidth]{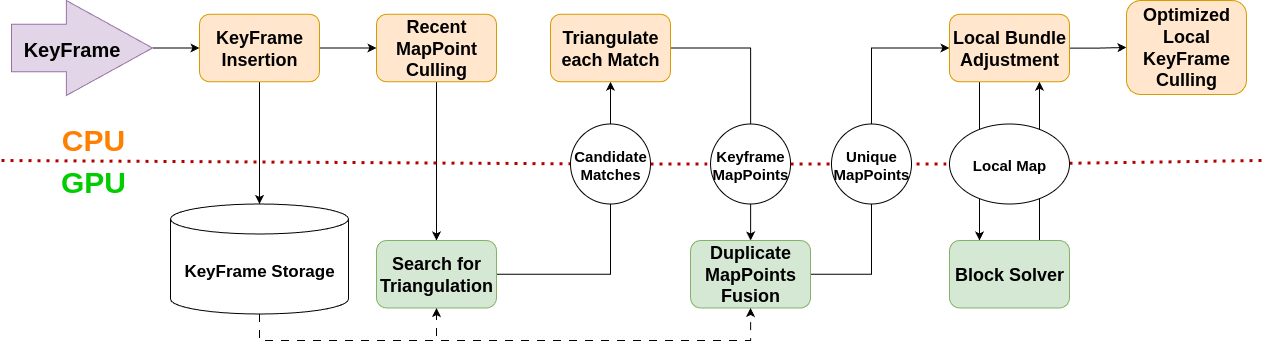}
    \caption{Data flow of the local mapping process in \SYS{}. CPU-side components are shown at the top, and GPU-side components at the bottom. Arrows indicate data transfers between modules. Keyframe storage resides on the GPU.}
    \label{fig:fastMap_breakdown}
    \vspace{-15pt}
\end{figure*}

\section{Design} \label{sec:design}

As shown in \autoref{fig:local_mapping_time_breakdown}, approximately 95\% of computation time in local mapping is spent on Map Point Creation, Map Point Fusion, Local Bundle Adjustment (LBA), and Keyframe Culling. For GPU-based parallelization, Map Point Creation, Map Point Fusion, and Keyframe Culling are challenging because they frequently update shared data structures, such as map points and keyframes, which requires synchronization to maintain consistency. These shared-state updates create serialization points and limit the benefits of GPU execution. In contrast, previous work has successfully accelerated LBA on the GPU because it is naturally parallelizable and involves minimal synchronization or shared-state updates. In addition to parallelization challenges, all of these components operate on large collections of keyframe and map point data, and repeatedly transferring such heavy data structures to the GPU can introduce significant data movement overhead, further limiting performance if not managed carefully.

To address these challenges, \SYS{} takes a holistic approach: we restructure computations, manage data efficiently, and apply parallel execution where feasible. Components are parallelized only if they incur high computational cost, allow the elimination of synchronization that would otherwise hinder parallel execution, and avoid repeated heavy data transfers between the CPU and GPU. In the following sections, we describe each component of \SYS{} in detail, explaining the underlying challenges and the corresponding solutions that enable efficient execution. Throughout this discussion, we refer to \autoref{fig:fastMap_breakdown}, which provides an overview of the local mapping architecture and illustrates the division of responsibilities between the CPU and GPU.

\subsection{Parallelization Challenges}
\label{design:challenges}

Parallelizing local mapping is constrained by two fundamental architectural challenges. First, the main stages in local mapping are tightly coupled with updates to shared map structures. During Map Point Creation, newly triangulated points are inserted into the global map and linked to observing keyframes. In Map Point Fusion, duplicate points are merged and their observations are reassigned. In Keyframe Culling, removing a redundant keyframe requires updating covisibility relationships and observation lists. In each of these stages, the values of keyframes or map points are modified, and these data structures are shared with other concurrent processes such as tracking and loop closing. Because multiple processes may access or modify the same objects, synchronization is necessary to preserve consistency. As a result, these interleaved updates create serialization points that limit parallel execution.

Second, local mapping operates on large and complex CPU-resident data structures, including keyframes and map points. Each keyframe contains hundreds or thousands of feature descriptors, geometric information, and observation links, while the map may consist of tens of thousands of points. Transferring such large amounts of data to the GPU at each iteration would require substantial bandwidth and memory operations, introducing significant runtime overhead. Moreover, since these structures are frequently updated, naive copying of data would not only be slow but also require careful synchronization to maintain correctness. This makes the size and mutability of the data a major barrier to efficient GPU-based parallelization.

We address the first challenge by restructuring the computations in Map Point Creation and moving the Keypoint Correspondence Search subroutine to the GPU, as well as restructuring the computations in the Fuse stage. To address the second challenge, we introduce a persistent keyframe storage on the GPU. In the following sections, we first describe the keyframe storage design, and then detail how each component addresses the challenge of synchronization-induced serialization.

\subsection{Keyframe Storage}
\label{design:keyframe_storage}

Keypoint Correspondence Search and Map Point Fusion, the two components we parallelize on the GPU, frequently require access to map point data from multiple neighboring keyframes. A naive approach would transfer all required keyframes to the GPU whenever a new keyframe is processed. Given that each keyframe occupies approximately 1 MB and that tens of neighboring keyframes may be involved, repeated transfers would introduce significant overhead.

To address this, \SYS{} introduces a persistent GPU keyframe storage. We copy each keyframe to the GPU only once when it enters local mapping and keep it resident until it is removed from the map or the system shuts down. To reduce both data transfer and synchronization overhead, we partition keyframe data into immutable and mutable components, storing only the immutable data on the GPU. Immutable data, such as keypoints and their descriptors, do not change after keyframe creation, so keeping them on the GPU eliminates the need for costly synchronization with the CPU. Mutable data, by contrast, can change during local mapping and would require frequent updates if kept on the GPU, which would severely reduce efficiency. We also exclude other keyframe data that are unnecessary for GPU computations. By keeping only the necessary immutable data on the GPU, we avoid repeated transfers and synchronization, and enable efficient parallel execution of parallelized stages.

\subsection{Keypoint Correspondence Search}
\label{design:search_for_triangulation}

As shown in \autoref{fig:fastMap_breakdown}, the first component we parallelize is the Keypoint Correspondence Search stage within Map Point Creation. This stage performs keypoint correspondence search between the current keyframe and its neighboring keyframes. For each keypoint in the current keyframe, the system identifies candidate matches in neighboring keyframes by computing descriptor distances and validating them using epipolar constraints. In the original non-parallel structure, neighboring keyframes are processed sequentially, and triangulation is performed immediately after each valid match is found. Because triangulation inserts new map points into the shared map and updates keyframe and map point structures, these operations require synchronization to preserve consistency. As a result, correspondence search and map updates are tightly interleaved, creating serialization points that prevent parallel execution and make GPU parallelization ineffective.

To address this limitation, we restructure the computation to enable parallel execution. We decouple Keypoint Correspondence Search from the Triangulation step, allowing matches across all neighboring keyframes to be identified before creating new map points and removing synchronization-dependent operations from the search stage. To handle data transfers efficiently, we take two measures. First, we eliminate repeated synchronization by batching the required map point data and transferring it to the GPU in a single step prior to parallel execution. Second, most of the data required at this stage originate from neighboring keyframes. Because these keyframes are already stored in persistent GPU memory as part of our keyframe storage design, no additional transfers are necessary. Each GPU thread then independently performs descriptor comparisons and epipolar validation, distributing the workload across thousands of threads. By combining structural code changes with GPU-resident data, we significantly reduce latency and enable efficient parallel execution.

\subsection{Map-point Fusion}
\label{design:fuse}

Duplicate Map Point Fusion is the second component \SYS{} parallelizes on the GPU as shown in \autoref{fig:fastMap_breakdown}. The primary goal of this stage is to determine whether there exists a map point in the map that (1) has descriptors very similar to those of a map point associated with the current keyframe and (2) is spatially close to that map point. This process is computationally intensive and has the potential to be parallelizable. It also benefits from our keyframe storage design, as most of the required data, particularly keyframe-related information, already reside in GPU memory.
However, similar to Keypoint Correspondence Search, the original structure is sequential, where the system loops over neighboring keyframes and their associated map points, and once a duplicate is found, it immediately fuses the map points. Since fusion modifies map point structures and requires synchronization, this tightly coupled logic prevents straightforward parallelization.

To enable GPU acceleration, we apply a restructuring strategy similar to that used in Keypoint Correspondence Search. Instead of fusing duplicates as they are found, we first determine all candidate duplicate pairs in parallel and defer the actual fusion to a subsequent sequential stage. In the GPU implementation, each thread is responsible for identifying the most similar map point to a given point in the current keyframe within a specific neighboring keyframe. To eliminate synchronization during the search phase, we gather all required map point data in a single batch and transfer it to the GPU for processing. The neighboring keyframe data are already stored on the GPU, incurring no transfer cost, while the map points associated with the current keyframe amount to approximately 100 KB, which has negligible impact on performance. By separating search from fusion and removing synchronization from the parallel stage, we enable efficient GPU execution while preserving the correctness of the original logic.

\subsection{Local Bundle Adjustment}
\label{design:LBA}
The final component that \SYS{} accelerates by using GPU computation is Local Bundle Adjustment, as illustrated in \autoref{fig:fastMap_breakdown}. As discussed in \autoref{sec:related-work}, several prior efforts have been made to accelerate this component. Since this component has been extensively studied and optimized in prior work, we chose to adopt an existing implementation rather than develop a new one from scratch. Specifically, we integrate the GPU-accelerated local bundle adjustment from the work of Gopinath et al.~\cite{Gopinath2023}, which is compatible with ORB-SLAM3. In this work, the \emph{Schur Complement} method, which is used to reduce the system size and speed up the LBA process, is parallelized on the GPU. Their implementation serves as a drop-in replacement block solver for g2o~\cite{Kummerle2011}, the optimization library used by ORB-SLAM3 to solve the local bundle adjustment problem. The GPU solver has been numerically validated against the original CPU-based g2o implementation, showing negligible differences in reprojection error, estimated states, and convergence behavior. It preserves the same mathematical formulation and optimization framework, ensuring that the acceleration improves efficiency without affecting accuracy or stability.

\subsection{Local Keyframe Culling}
\label{design:kf_culling}
Local Keyframe Culling is the final major component of local mapping. We do not parallelize this stage, mainly due to the significant data transfer overhead it would have. Instead, we apply a CPU-side optimization to improve its performance while avoiding the cost of GPU data transfers.

As described in \autoref{sec:background}, the Keyframe Culling stage removes redundant keyframes by checking how many of their associated map points are also observed by other keyframes in the map at the same or finer scale, and discarding those that exceed a predefined threshold. To perform this check, the system maintains an observation list for each map point, which records the set of keyframes that observe that point. This data structure is large and frequently updated during each iteration of local mapping. Unlike the relatively static keyframe data which can be stored on the GPU without frequent synchronization, the dynamic nature of the observation data makes it unsuitable for the same treatment. We experimented with a GPU-side map point storage system that continuously updated these data, but found that maintaining consistency introduced significant overhead. In fact, the cost of keeping this storage synchronized was nearly equivalent to the total cost of the entire local mapping stage, making the approach impractical. As a result, we chose not to parallelize this component on the GPU.

Instead of parallelizing this logic on the GPU, we optimize it on the CPU by introducing a lightweight array that tracks how many times each map point is observed at each scale. Each time a map point is observed, the corresponding counter in the array is incremented. This approach eliminates the need to iterate over every keyframe's observations to count the number of valid observations per map point. As a result, it significantly reduces the computational overhead of the keyframe culling process. The additional memory overhead of this optimization is minimal, as we store only a small array per map point, which amounts to just a few bytes.

\begin{table*}[ht]
  \centering
  \begin{subtable}{\textwidth}
    \centering
    \renewcommand{\arraystretch}{1.4}
\vspace{10pt}
\centering
\resizebox{1\linewidth}{!}{
\setlength{\tabcolsep}{4pt}
\begin{tabular}{|c|
*{2}{>{\centering\arraybackslash}p{1.5cm}|}p{0.7cm}|
*{2}{>{\centering\arraybackslash}p{0.7cm}|}p{0.7cm}|
*{2}{>{\centering\arraybackslash}p{0.7cm}|}p{0.7cm}|
*{2}{>{\centering\arraybackslash}p{0.7cm}|}p{0.7cm}|
*{2}{>{\centering\arraybackslash}p{0.7cm}|}p{0.7cm}|
*{2}{>{\centering\arraybackslash}p{0.7cm}|}}
\hline
\multirow{2}{*}{\textbf{Sequence}} &
\multicolumn{2}{c|}{\textbf{Total Local Mapping}} & \multicolumn{1}{c|}{\textbf{Speed}} &
\multicolumn{2}{c|}{\textbf{KP Search}} & \multicolumn{1}{c|}{\textbf{Speed}} &
\multicolumn{2}{c|}{\textbf{MP Fusion}} & \multicolumn{1}{c|}{\textbf{Speed}} &
\multicolumn{2}{c|}{\textbf{LBA}} & \multicolumn{1}{c|}{\textbf{Speed}} &
\multicolumn{2}{c|}{\textbf{KF Culling}} & \multicolumn{1}{c|}{\textbf{Speed}} &
\multicolumn{2}{c|}{\textbf{ATE (m)}} \\
\cline{2-3} \cline{5-6} \cline{8-9} \cline{11-12} \cline{14-15} \cline{17-18}
 & \textbf{Original} & \textbf{\SYS{}} & \multicolumn{1}{c|}{\textbf{Up}} 
 & \textbf{Org} & \textbf{TM} & \multicolumn{1}{c|}{\textbf{Up}} 
 & \textbf{Org} & \textbf{TM} & \multicolumn{1}{c|}{\textbf{Up}} 
 & \textbf{Org} & \textbf{TM} & \multicolumn{1}{c|}{\textbf{Up}}
 & \textbf{Org} & \textbf{TM} & \multicolumn{1}{c|}{\textbf{Up}}
 & \textbf{Org} & \textbf{TM} \\
\hline
\textbf{EuRoC Avg} & 372.70 & 287.90 & \textcolor{Green}{\textbf{\hspace{+0.09cm}{1.3x}}} & 7.10 & 2.04 & \textcolor{Green}{\textbf{\hspace{+0.09cm}{3.5x}}} & 38.06 & 27.72 & \textcolor{Green}{\textbf{\hspace{+0.09cm}{1.4x}}} & 276.37 & 218.28 & \textcolor{Green}{\textbf{\hspace{+0.09cm}{1.3x}}} & 31.60 & 17.45 & \textcolor{Green}{\textbf{\hspace{+0.09cm}{1.8x}}} & 0.034 & 0.031 \\
\hline
\textbf{MH01} & $364\pm139$ & $278\pm104$ & \textcolor{Green}{\textbf{\hspace{+0.09cm}{1.3x}}} & $8.6$ & $2.1$ & \textcolor{Green}{\textbf{\hspace{+0.09cm}{4.1x}}} & $37.6$ & $25.9$ & \textcolor{Green}{\textbf{\hspace{+0.09cm}{1.5x}}} & $267.4$ & $209.6$ & \textcolor{Green}{\textbf{\hspace{+0.09cm}{1.3x}}} & $29.7$ & $17.4$ & \textcolor{Green}{\textbf{\hspace{+0.09cm}{1.7x}}} & $0.040$ & $0.039$ \\
\hline
\textbf{MH02} & $322\pm127$ & $249\pm95$ & \textcolor{Green}{\textbf{\hspace{+0.09cm}{1.3x}}} & $8.0$ & $2.2$ & \textcolor{Green}{\textbf{\hspace{+0.09cm}{3.7x}}} & $33.2$ & $22.7$ & \textcolor{Green}{\textbf{\hspace{+0.09cm}{1.5x}}} & $238.7$ & $188.8$ & \textcolor{Green}{\textbf{\hspace{+0.09cm}{1.3x}}} & $23.9$ & $14.1$ & \textcolor{Green}{\textbf{\hspace{+0.09cm}{1.7x}}} & $0.032$ & $0.032$ \\
\hline
\textbf{V101} & $425\pm101$ & $321\pm75$ & \textcolor{Green}{\textbf{\hspace{+0.09cm}{1.3x}}} & $5.7$ & $1.9$ & \textcolor{Green}{\textbf{\hspace{+0.09cm}{3.0x}}} & $43.1$ & $31.7$ & \textcolor{Green}{\textbf{\hspace{+0.09cm}{1.4x}}} & $313.8$ & $243.9$ & \textcolor{Green}{\textbf{\hspace{+0.09cm}{1.3x}}} & $42.0$ & $20.6$ & \textcolor{Green}{\textbf{\hspace{+0.09cm}{2.0x}}} & $0.038$ & $0.038$ \\
\hline
\textbf{V102} & $379\pm116$ & $302\pm95$ & \textcolor{Green}{\textbf{\hspace{+0.09cm}{1.3x}}} & $6.1$ & $2.0$ & \textcolor{Green}{\textbf{\hspace{+0.09cm}{3.1x}}} & $38.4$ & $30.5$ & \textcolor{Green}{\textbf{\hspace{+0.09cm}{1.3x}}} & $285.6$ & $230.8$ & \textcolor{Green}{\textbf{\hspace{+0.09cm}{1.2x}}} & $30.7$ & $17.7$ & \textcolor{Green}{\textbf{\hspace{+0.09cm}{1.7x}}} & $0.027$ & $0.018$ \\
\Xhline{4\arrayrulewidth}
\textbf{TUM-VI Avg} & 551.63 & 356.85 & \textcolor{Green}{\textbf{\hspace{+0.09cm}{1.6x}}} & 33.68 & 6.45 & \textcolor{Green}{\textbf{\hspace{+0.09cm}{5.2x}}} & 90.42 & 40.79 & \textcolor{Green}{\textbf{\hspace{+0.09cm}{2.2x}}} & 342.99 & 257.45 & \textcolor{Green}{\textbf{\hspace{+0.09cm}{1.3x}}} & 55.49 & 17.43 & \textcolor{Green}{\textbf{\hspace{+0.09cm}{3.2x}}} & 0.013 & 0.013 \\
\hline
\textbf{room1} & $584\pm164$ & $387\pm103$ & \textcolor{Green}{\textbf{\hspace{+0.09cm}{1.5x}}} & $30.3$ & $5.8$ & \textcolor{Green}{\textbf{\hspace{+0.09cm}{5.3x}}} & $106.1$ & $48.2$ & \textcolor{Green}{\textbf{\hspace{+0.09cm}{2.2x}}} & $356.8$ & $278.6$ & \textcolor{Green}{\textbf{\hspace{+0.09cm}{1.3x}}} & $59.0$ & $16.3$ & \textcolor{Green}{\textbf{\hspace{+0.09cm}{3.6x}}} & $0.008$ & $0.009$ \\
\hline
\textbf{room2} & $529\pm159$ & $364\pm118$ & \textcolor{Green}{\textbf{\hspace{+0.09cm}{1.5x}}} & $29.8$ & $5.6$ & \textcolor{Green}{\textbf{\hspace{+0.09cm}{5.4x}}} & $89.3$ & $42.2$ & \textcolor{Green}{\textbf{\hspace{+0.09cm}{2.1x}}} & $337.0$ & $265.4$ & \textcolor{Green}{\textbf{\hspace{+0.09cm}{1.3x}}} & $43.6$ & $14.2$ & \textcolor{Green}{\textbf{\hspace{+0.09cm}{3.1x}}} & $0.011$ & $0.012$ \\
\hline
\textbf{corridor1} & $546\pm126$ & $339\pm63$ & \textcolor{Green}{\textbf{\hspace{+0.09cm}{1.6x}}} & $36.5$ & $7.2$ & \textcolor{Green}{\textbf{\hspace{+0.09cm}{5.1x}}} & $82.6$ & $36.4$ & \textcolor{Green}{\textbf{\hspace{+0.09cm}{2.3x}}} & $339.8$ & $244.2$ & \textcolor{Green}{\textbf{\hspace{+0.09cm}{1.4x}}} & $60.9$ & $19.8$ & \textcolor{Green}{\textbf{\hspace{+0.09cm}{3.1x}}} & $0.015$ & $0.017$ \\
\hline
\textbf{corridor2} & $545\pm112$ & $335\pm57$ & \textcolor{Green}{\textbf{\hspace{+0.09cm}{1.6x}}} & $38.1$ & $7.3$ & \textcolor{Green}{\textbf{\hspace{+0.09cm}{5.2x}}} & $83.7$ & $36.4$ & \textcolor{Green}{\textbf{\hspace{+0.09cm}{2.3x}}} & $338.3$ & $241.6$ & \textcolor{Green}{\textbf{\hspace{+0.09cm}{1.4x}}} & $58.4$ & $19.5$ & \textcolor{Green}{\textbf{\hspace{+0.09cm}{3.0x}}} & $0.020$ & $0.018$ \\
\hline
\end{tabular}
}
\label{tab:jetson_runtime_comparison}
    \caption{Jetson Results}
    \label{tab:jetson}
  \end{subtable}
  
  \begin{subtable}{\textwidth}
    \centering
    \renewcommand{\arraystretch}{1.4}
\vspace{10pt}
\centering
\resizebox{1\linewidth}{!}{
\setlength{\tabcolsep}{4pt}
\begin{tabular}{|c|
*{2}{>{\centering\arraybackslash}p{1.5cm}|}p{0.7cm}|
*{2}{>{\centering\arraybackslash}p{0.7cm}|}p{0.7cm}|
*{2}{>{\centering\arraybackslash}p{0.7cm}|}p{0.7cm}|
*{2}{>{\centering\arraybackslash}p{0.7cm}|}p{0.7cm}|
*{2}{>{\centering\arraybackslash}p{0.7cm}|}p{0.7cm}|
*{2}{>{\centering\arraybackslash}p{0.7cm}|}}
\hline
\multirow{2}{*}{\textbf{Sequence}} &
\multicolumn{2}{c|}{\textbf{Total Local Mapping}} & \multicolumn{1}{c|}{\textbf{Speed}} &
\multicolumn{2}{c|}{\textbf{KP Search}} & \multicolumn{1}{c|}{\textbf{Speed}} &
\multicolumn{2}{c|}{\textbf{MP Fusion}} & \multicolumn{1}{c|}{\textbf{Speed}} &
\multicolumn{2}{c|}{\textbf{LBA}} & \multicolumn{1}{c|}{\textbf{Speed}} &
\multicolumn{2}{c|}{\textbf{KF Culling}} & \multicolumn{1}{c|}{\textbf{Speed}} &
\multicolumn{2}{c|}{\textbf{ATE (m)}} \\
\cline{2-3} \cline{5-6} \cline{8-9} \cline{11-12} \cline{14-15} \cline{17-18}
 & \textbf{Original} & \textbf{\SYS{}} & \multicolumn{1}{c|}{\textbf{Up}} 
 & \textbf{Org} & \textbf{TM} & \multicolumn{1}{c|}{\textbf{Up}} 
 & \textbf{Org} & \textbf{TM} & \multicolumn{1}{c|}{\textbf{Up}} 
 & \textbf{Org} & \textbf{TM} & \multicolumn{1}{c|}{\textbf{Up}}
 & \textbf{Org} & \textbf{TM} & \multicolumn{1}{c|}{\textbf{Up}}
 & \textbf{Org} & \textbf{TM} \\
\hline
\textbf{EuRoC Avg} & 65.73 & 51.56 & \textcolor{Green}{\textbf{\hspace{+0.09cm}{1.3x}}} & 1.60 & 0.73 & \textcolor{Green}{\textbf{\hspace{+0.09cm}{2.2x}}} & 6.83 & 4.60 & \textcolor{Green}{\textbf{\hspace{+0.09cm}{1.5x}}} & 48.52 & 39.05 & \textcolor{Green}{\textbf{\hspace{+0.09cm}{1.2x}}} & 4.45 & 2.42 & \textcolor{Green}{\textbf{\hspace{+0.09cm}{1.8x}}} & 0.036 & 0.034 \\
\hline
\textbf{MH01} & $65.0\pm26.4$ & $49.2\pm18.7$ & \textcolor{Green}{\textbf{\hspace{+0.09cm}{1.3x}}} & $1.8$ & $0.7$ & \textcolor{Green}{\textbf{\hspace{+0.09cm}{2.6x}}} & $6.7$ & $4.3$ & \textcolor{Green}{\textbf{\hspace{+0.09cm}{1.6x}}} & $47.0$ & $36.8$ & \textcolor{Green}{\textbf{\hspace{+0.09cm}{1.3x}}} & $4.9$ & $2.6$ & \textcolor{Green}{\textbf{\hspace{+0.09cm}{1.9x}}} & $0.042$ & $0.036$ \\
\hline
\textbf{MH02} & $58.7\pm24.1$ & $45.1\pm17.3$ & \textcolor{Green}{\textbf{\hspace{+0.09cm}{1.3x}}} & $1.8$ & $0.7$ & \textcolor{Green}{\textbf{\hspace{+0.09cm}{2.5x}}} & $6.1$ & $3.8$ & \textcolor{Green}{\textbf{\hspace{+0.09cm}{1.6x}}} & $42.7$ & $33.9$ & \textcolor{Green}{\textbf{\hspace{+0.09cm}{1.3x}}} & $3.9$ & $2.1$ & \textcolor{Green}{\textbf{\hspace{+0.09cm}{1.8x}}} & $0.032$ & $0.031$ \\
\hline
\textbf{MH03} & $70.1\pm25.3$ & $54.1\pm19.4$ & \textcolor{Green}{\textbf{\hspace{+0.09cm}{1.3x}}} & $1.7$ & $0.7$ & \textcolor{Green}{\textbf{\hspace{+0.09cm}{2.4x}}} & $7.4$ & $4.7$ & \textcolor{Green}{\textbf{\hspace{+0.09cm}{1.6x}}} & $51.9$ & $41.3$ & \textcolor{Green}{\textbf{\hspace{+0.09cm}{1.3x}}} & $4.6$ & $2.6$ & \textcolor{Green}{\textbf{\hspace{+0.09cm}{1.8x}}} & $0.034$ & $0.030$ \\
\hline
\textbf{MH04} & $63.3\pm20.1$ & $51.8\pm15.2$ & \textcolor{Green}{\textbf{\hspace{+0.09cm}{1.2x}}} & $1.8$ & $1.0$ & \textcolor{Green}{\textbf{\hspace{+0.09cm}{1.9x}}} & $6.3$ & $4.6$ & \textcolor{Green}{\textbf{\hspace{+0.09cm}{1.4x}}} & $47.3$ & $38.7$ & \textcolor{Green}{\textbf{\hspace{+0.09cm}{1.2x}}} & $3.4$ & $2.3$ & \textcolor{Green}{\textbf{\hspace{+0.09cm}{1.5x}}} & $0.045$ & $0.040$ \\
\hline
\textbf{MH05} & $63.8\pm19.5$ & $51.4\pm14.9$ & \textcolor{Green}{\textbf{\hspace{+0.09cm}{1.2x}}} & $1.8$ & $1.0$ & \textcolor{Green}{\textbf{\hspace{+0.09cm}{1.8x}}} & $6.4$ & $4.5$ & \textcolor{Green}{\textbf{\hspace{+0.09cm}{1.4x}}} & $47.6$ & $38.4$ & \textcolor{Green}{\textbf{\hspace{+0.09cm}{1.2x}}} & $3.4$ & $2.2$ & \textcolor{Green}{\textbf{\hspace{+0.09cm}{1.5x}}} & $0.054$ & $0.054$ \\
\hline
\textbf{V101} & $77.2\pm18.8$ & $58.4\pm14.1$ & \textcolor{Green}{\textbf{\hspace{+0.09cm}{1.3x}}} & $1.3$ & $0.5$ & \textcolor{Green}{\textbf{\hspace{+0.09cm}{2.4x}}} & $8.0$ & $5.4$ & \textcolor{Green}{\textbf{\hspace{+0.09cm}{1.5x}}} & $56.6$ & $44.6$ & \textcolor{Green}{\textbf{\hspace{+0.09cm}{1.3x}}} & $6.7$ & $3.0$ & \textcolor{Green}{\textbf{\hspace{+0.09cm}{2.3x}}} & $0.038$ & $0.038$ \\
\hline
\textbf{V102} & $69.7\pm22.6$ & $55.5\pm17.7$ & \textcolor{Green}{\textbf{\hspace{+0.09cm}{1.3x}}} & $1.4$ & $0.6$ & \textcolor{Green}{\textbf{\hspace{+0.09cm}{2.4x}}} & $7.2$ & $5.1$ & \textcolor{Green}{\textbf{\hspace{+0.09cm}{1.4x}}} & $52.2$ & $42.8$ & \textcolor{Green}{\textbf{\hspace{+0.09cm}{1.2x}}} & $4.9$ & $2.6$ & \textcolor{Green}{\textbf{\hspace{+0.09cm}{1.9x}}} & $0.018$ & $0.018$ \\
\hline
\textbf{V103} & $58.0\pm15.5$ & $46.9\pm11.7$ & \textcolor{Green}{\textbf{\hspace{+0.09cm}{1.2x}}} & $1.2$ & $0.6$ & \textcolor{Green}{\textbf{\hspace{+0.09cm}{2.1x}}} & $6.5$ & $4.5$ & \textcolor{Green}{\textbf{\hspace{+0.09cm}{1.5x}}} & $43.0$ & $35.9$ & \textcolor{Green}{\textbf{\hspace{+0.09cm}{1.2x}}} & $3.9$ & $2.1$ & \textcolor{Green}{\textbf{\hspace{+0.09cm}{1.8x}}} & $0.026$ & $0.026$ \\
\Xhline{4\arrayrulewidth}
\textbf{TUM-VI Avg} & 109.81 & 69.84 & \textcolor{Green}{\textbf{\hspace{+0.09cm}{1.6x}}} & 7.68 & 2.36 & \textcolor{Green}{\textbf{\hspace{+0.09cm}{3.3x}}} &17.88 & 7.63 & \textcolor{Green}{\textbf{\hspace{+0.09cm}{2.3x}}} & 67.06 & 49.10 & \textcolor{Green}{\textbf{\hspace{+0.09cm}{1.4x}}} &10.43 & 3.16 & \textcolor{Green}{\textbf{\hspace{+0.09cm}{3.3x}}} & 0.012 & 0.013 \\
\hline
\textbf{room1} & $113.2\pm33.8$ & $75.1\pm19.6$ & \textcolor{Green}{\textbf{\hspace{+0.09cm}{1.5x}}} & $7.1$ & $2.1$ & \textcolor{Green}{\textbf{\hspace{+0.09cm}{3.3x}}} & $19.8$ & $9.0$ & \textcolor{Green}{\textbf{\hspace{+0.09cm}{2.2x}}} & $68.9$ & $52.8$ & \textcolor{Green}{\textbf{\hspace{+0.09cm}{1.3x}}} & $10.3$ & $3.0$ & \textcolor{Green}{\textbf{\hspace{+0.09cm}{3.5x}}} & $0.009$ & $0.013$ \\
\hline
\textbf{room2} & $103.2\pm32.3$ & $72.6\pm20.3$ & \textcolor{Green}{\textbf{\hspace{+0.09cm}{1.4x}}} & $7.2$ & $2.0$ & \textcolor{Green}{\textbf{\hspace{+0.09cm}{3.5x}}} & $16.9$ & $7.8$ & \textcolor{Green}{\textbf{\hspace{+0.09cm}{2.1x}}} & $64.6$ & $52.1$ & \textcolor{Green}{\textbf{\hspace{+0.09cm}{1.2x}}} & $7.9$ & $2.7$ & \textcolor{Green}{\textbf{\hspace{+0.09cm}{3.0x}}} & $0.012$ & $0.012$ \\
\hline
\textbf{room3} & $111.5\pm29.4$ & $69.4\pm14.6$ & \textcolor{Green}{\textbf{\hspace{+0.09cm}{1.6x}}} & $7.3$ & $2.1$ & \textcolor{Green}{\textbf{\hspace{+0.09cm}{3.5x}}} & $16.9$ & $6.9$ & \textcolor{Green}{\textbf{\hspace{+0.09cm}{2.5x}}} & $73.1$ & $50.7$ & \textcolor{Green}{\textbf{\hspace{+0.09cm}{1.4x}}} & $7.8$ & $2.5$ & \textcolor{Green}{\textbf{\hspace{+0.09cm}{3.2x}}} & $0.009$ & $0.014$ \\
\hline
\textbf{room4} & $101.9\pm32.4$ & $64.1\pm14.3$ & \textcolor{Green}{\textbf{\hspace{+0.09cm}{1.6x}}} & $7.0$ & $2.2$ & \textcolor{Green}{\textbf{\hspace{+0.09cm}{3.2x}}} & $18.1$ & $7.3$ & \textcolor{Green}{\textbf{\hspace{+0.09cm}{2.5x}}} & $60.7$ & $44.7$ & \textcolor{Green}{\textbf{\hspace{+0.09cm}{1.4x}}} & $9.1$ & $2.4$ & \textcolor{Green}{\textbf{\hspace{+0.09cm}{3.7x}}} & $0.009$ & $0.009$ \\
\hline
\textbf{room5} & $112.6\pm31.1$ & $74.4\pm17.4$ & \textcolor{Green}{\textbf{\hspace{+0.09cm}{1.5x}}} & $7.3$ & $2.3$ & \textcolor{Green}{\textbf{\hspace{+0.09cm}{3.3x}}} & $19.7$ & $8.8$ & \textcolor{Green}{\textbf{\hspace{+0.09cm}{2.3x}}} & $68.1$ & $52.3$ & \textcolor{Green}{\textbf{\hspace{+0.09cm}{1.3x}}} & $10.3$ & $3.0$ & \textcolor{Green}{\textbf{\hspace{+0.09cm}{3.5x}}} & $0.011$ & $0.012$ \\
\hline
\textbf{room6} & $114.7\pm31.0$ & $71.5\pm16.6$ & \textcolor{Green}{\textbf{\hspace{+0.09cm}{1.6x}}} & $6.8$ & $2.1$ & \textcolor{Green}{\textbf{\hspace{+0.09cm}{3.2x}}} & $21.6$ & $9.8$ & \textcolor{Green}{\textbf{\hspace{+0.09cm}{2.2x}}} & $67.5$ & $48.0$ & \textcolor{Green}{\textbf{\hspace{+0.09cm}{1.4x}}} & $10.9$ & $2.7$ & \textcolor{Green}{\textbf{\hspace{+0.09cm}{4.0x}}} & $0.006$ & $0.006$ \\
\hline
\textbf{corridor1} & $110.6\pm28.8$ & $67.8\pm12.3$ & \textcolor{Green}{\textbf{\hspace{+0.09cm}{1.6x}}} & $8.6$ & $2.8$ & \textcolor{Green}{\textbf{\hspace{+0.09cm}{3.1x}}} & $15.8$ & $6.3$ & \textcolor{Green}{\textbf{\hspace{+0.09cm}{2.5x}}} & $67.4$ & $47.7$ & \textcolor{Green}{\textbf{\hspace{+0.09cm}{1.4x}}} & $12.5$ & $4.1$ & \textcolor{Green}{\textbf{\hspace{+0.09cm}{3.0x}}} & $0.020$ & $0.021$ \\
\hline
\textbf{corridor2} & $110.3\pm25.5$ & $66.5\pm10.4$ & \textcolor{Green}{\textbf{\hspace{+0.09cm}{1.7x}}} & $8.9$ & $2.9$ & \textcolor{Green}{\textbf{\hspace{+0.09cm}{3.1x}}} & $15.9$ & $6.2$ & \textcolor{Green}{\textbf{\hspace{+0.09cm}{2.6x}}} & $67.2$ & $46.7$ & \textcolor{Green}{\textbf{\hspace{+0.09cm}{1.4x}}} & $12.2$ & $4.0$ & \textcolor{Green}{\textbf{\hspace{+0.09cm}{3.1x}}} & $0.019$ & $0.016$ \\
\hline
\textbf{corridor3} & $110.3\pm28.3$ & $67.0\pm11.5$ & \textcolor{Green}{\textbf{\hspace{+0.09cm}{1.6x}}} & $8.9$ & $2.8$ & \textcolor{Green}{\textbf{\hspace{+0.09cm}{3.2x}}} & $16.1$ & $6.5$ & \textcolor{Green}{\textbf{\hspace{+0.09cm}{2.5x}}} & $66.1$ & $46.8$ & \textcolor{Green}{\textbf{\hspace{+0.09cm}{1.4x}}} & $12.9$ & $4.2$ & \textcolor{Green}{\textbf{\hspace{+0.09cm}{3.1x}}} & $0.009$ & $0.014$ \\
\hline
\end{tabular}
}
\label{tab:desktop_runtime_comparison}
    \caption{Desktop Results}
    \label{tab:desktop}
  \end{subtable}

  \caption{Performance comparison of \SYS{} and ORB‑SLAM3 using the desktop and Jetson setups on \EUROC{} and \TUMVI{} datasets. We report mean$\pm$std runtime (ms), speed‑up, and ATE RMSE (m).}
  \label{tab:combined_results}
  \vspace{-10pt}
\end{table*}

\section{Evaluation} \label{sec:evaluation}

\begin{figure}[t]
    \centering
    \includegraphics[width=0.9\linewidth]{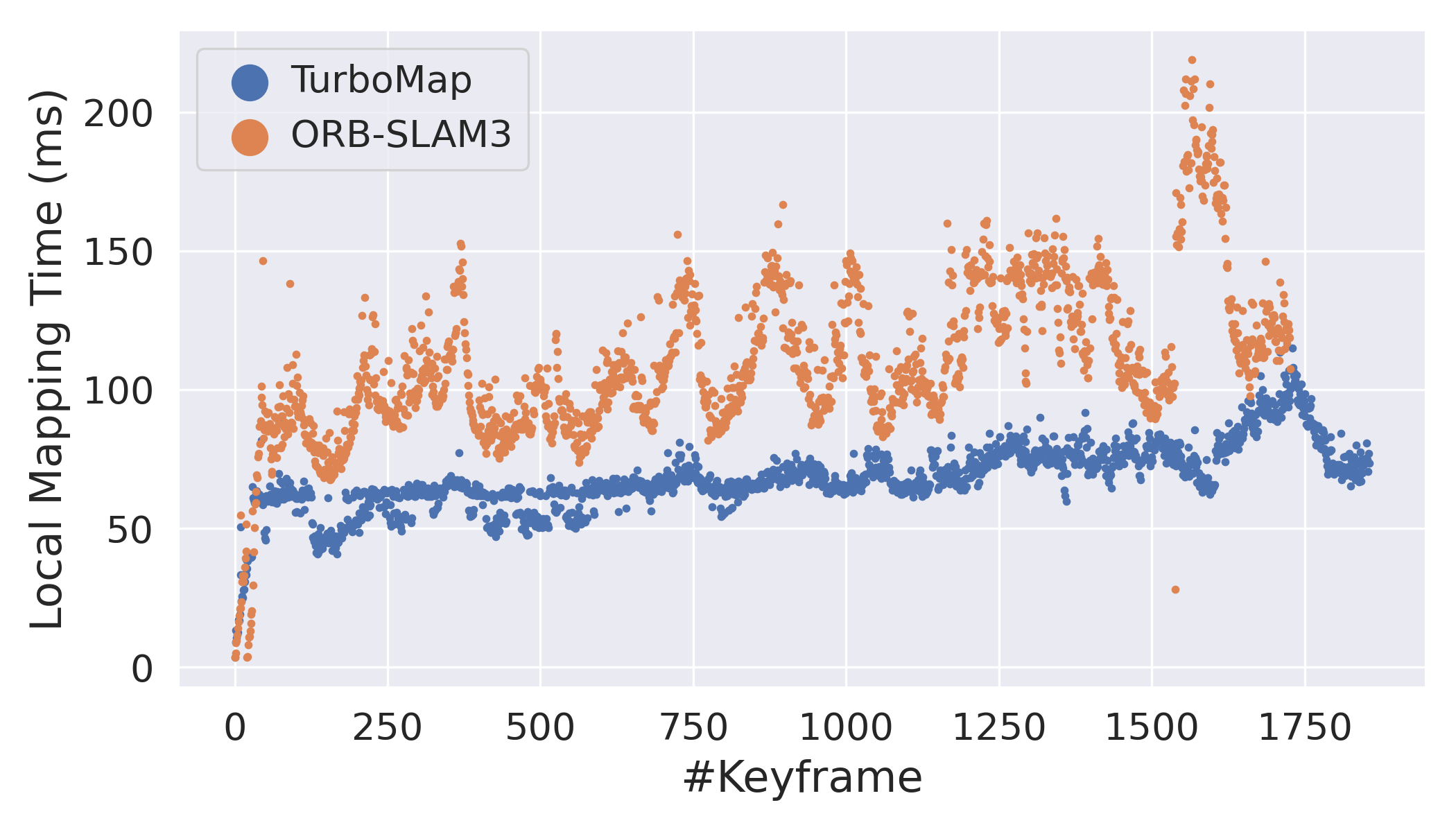}
    \caption{Corridor1 keyframe processing time for \SYS{} and ORB-SLAM3 in Desktop environment.}
    \label{fig:room1_time_comparison}
    \vspace{-15pt}
\end{figure}

\subsection{Experimental Setup}
\label{eval:setup}
We evaluate the performance of \SYS{} by running a set of sequences on two different hardware platforms and comparing the results with those of the original ORB-SLAM3. The experiments are conducted in two setups: a desktop machine and a mobile embedded system. The desktop system is equipped with a 20-core Intel Core i7-12700K CPU running at 5.0 GHz, an NVIDIA RTX 3090 GPU with 10,496 CUDA cores, and 64 GB of RAM. The embedded platform is a Jetson Xavier NX, featuring a 6-core ARM Carmel CPU running at 1.4 GHz, a 384-core NVIDIA Volta GPU, and 8 GB of RAM. To ensure consistent and reproducible results, we lock the CPU and GPU frequencies to their maximum values during all experiments.

We use a combination of sequences from the \TUMVI{} and \EUROC{} datasets, both in the stereo-inertial configuration. We run each sequence five times and report the average results across these runs. 

\subsection{Overall Performance}
\label{eval:overall_performance}
\autoref{tab:combined_results} presents the performance comparison between \SYS{} and ORB-SLAM3 for local mapping. The reported timings for each component include CPU–GPU data transfer overhead. We also conduct an ablation study where each accelerated component is enabled independently. The results show that each component operates modularly, and activating any individual optimization yields consistent improvements. As shown, \SYS{} gets an average speedup of $1.3\times$ in \EUROC{} and $1.6\times$ in \TUMVI{}, in both the desktop and embedded settings, when all optimizations are activated. Importantly, \SYS{} achieves these performance gains while maintaining a similar Absolute Trajectory Error (ATE) RMSE as ORB-SLAM3, demonstrating that our optimizations do not compromise the accuracy of the system.

In addition to improving the average processing speed, \SYS{} also significantly reduces the standard deviation of keyframe processing times. This improvement indicates that keyframes are processed with more consistent latency, reducing the occurrence of outlier frames that require significantly more time to process. As a result, fewer keyframes are at risk of being dropped due to excessive processing delays. 

To illustrate the impact of reduced variance, \autoref{fig:room1_time_comparison} shows the keyframe processing time for both systems in the Corridor1 sequence. As the number of keyframes in the map increases, ORB-SLAM3 exhibits a noticeable growth in processing time per keyframe. This is due to the increased workload across components such as triangulation, fusion, and local bundle adjustment, all of which scale with map size. In contrast, \SYS{} maintains more stable processing times even as the map grows, thanks to its GPU-accelerated components and CPU-side optimizations that effectively distribute and manage the computational load. This effect is clearly visible when comparing the maximum keyframe processing times in both systems: approximately 220 ms for ORB-SLAM3 versus only 115 ms for \SYS{}.

\subsection{Individual Component Performances}
\label{eval:individual_performance}
We analyze the performance gains from each optimization introduced in \SYS{} in the following subsections. It is important to note that the observed speedups differ between the \EUROC{} and \TUMVI{} datasets. This discrepancy arises from the different camera models and the way they are processed. Specifically, in \TUMVI{}, which employs fisheye cameras, the system does not rectify the images. Instead, it treats the stereo pair as two independent monocular cameras in order to leverage the wider field of view and retain more detected features. As a result, \TUMVI{} sequences contain approximately twice as many map points as those in \EUROC{}. Consequently, local mapping components, particularly those that scale with the number of map points, require significantly more processing time in \TUMVI{}.

\subsubsection{Keypoint Correspondence Search}
The GPU parallelization of the Keypoint Correspondence Search component results in an average speedup of $2.2\times$ on \EUROC{} and $3.3\times$ on \TUMVI{} in the desktop setting. In the Jetson setting, it achieves an average speedup of $3.5\times$ on \EUROC{} and $5.2\times$ on \TUMVI{}. The higher speedup achieved in \TUMVI{} is because the system is searching for more keypoint matches to triangulate in this dataset as we explained earlier, and \SYS{} uses more GPU threads in this optimization compared to \EUROC{}, resulting in greater performance gains.

\subsubsection{Map Point Fusion}
Parallelizing Map Point Fusion yields an average speedup of $1.5\times$ on \EUROC{} and $2.3\times$ on \TUMVI{} in the desktop setting. On the Jetson platform, it achieves average speedups of $1.4\times$ and $2.2\times$ on \EUROC{} and \TUMVI{}, respectively. The improved performance on \TUMVI{} is attributed to the higher number of map points being fused, allowing \SYS{} to exploit parallelism more effectively with a greater number of GPU threads.

\subsubsection{Keyframe Culling} 
\SYS{} optimizes the Keyframe Culling process on the CPU, achieving a speedup of $1.8\times$ on \EUROC{} on both desktop and Jetson. In \TUMVI{}, it achieves a speedup of $3.3\times$ on desktop and $3.2\times$ on Jetson.

\begin{figure}[t]
    \includegraphics[width=0.9\linewidth]{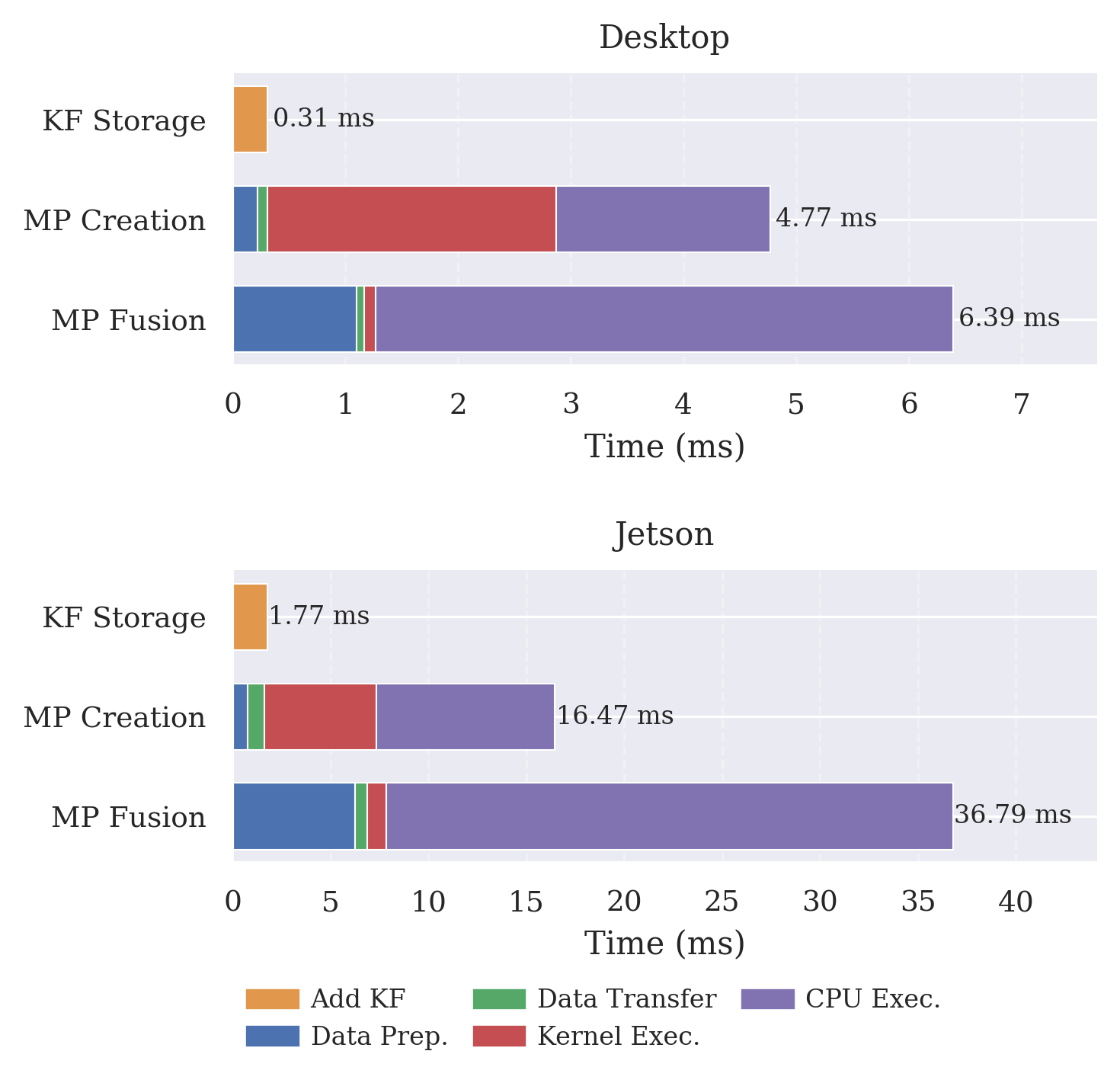}
    \caption{Timing breakdown of Map Point Creation and Fusion stages, with keyframe storage, in \SYS{} on Corridor1.}
    \label{fig:kernel_breakdown}
    \vspace{-15pt}
\end{figure}

\subsection{Map Point Creation and Fusion Timing Analysis}
\label{eval:kernel_breakdown}

\autoref{fig:kernel_breakdown} shows the average execution time of different stages of Map Point Creation and Fusion in \SYS{} on Corridor1, evaluated on Desktop and Jetson platforms, along with Keyframe Storage timings. CPU execution time corresponds to the portions remaining on the CPU due to negligible latency or unavoidable synchronization. Data preparation time reflects the cost of converting CPU-resident data into GPU-compatible formats and vice versa. This overhead is higher in Fusion, since selected map points must be restructured before transfer, whereas Map Point Creation primarily uses keyframe data already stored on the GPU.

In Map Point Creation, kernel execution dominates, indicating a computation-bound workload. In Fusion, data preparation dominates, reflecting the cost of restructuring map point data. Although transferring keyframes to the GPU is the most expensive data movement step, it is performed only once per local mapping iteration, preventing repeated transfer overhead. Importantly, the size of CPU–GPU data transfers does not grow with the size of the map. 
% The only data structure that scales with the map is keyframe data, which are transferred one at a time, and then kept resident on the GPU. As a result, per-iteration transfer costs remain stable even as the map expands.
The only data structure that grows with the map is the set of keyframes. In our design, keyframes are transferred to the GPU one at a time as they are created and then kept resident, avoiding repeated bulk transfers. As a result, the per-iteration data transfer cost remains stable as the map expands.

\begin{figure}[t]
    \centering
    \includegraphics[width=1\linewidth]{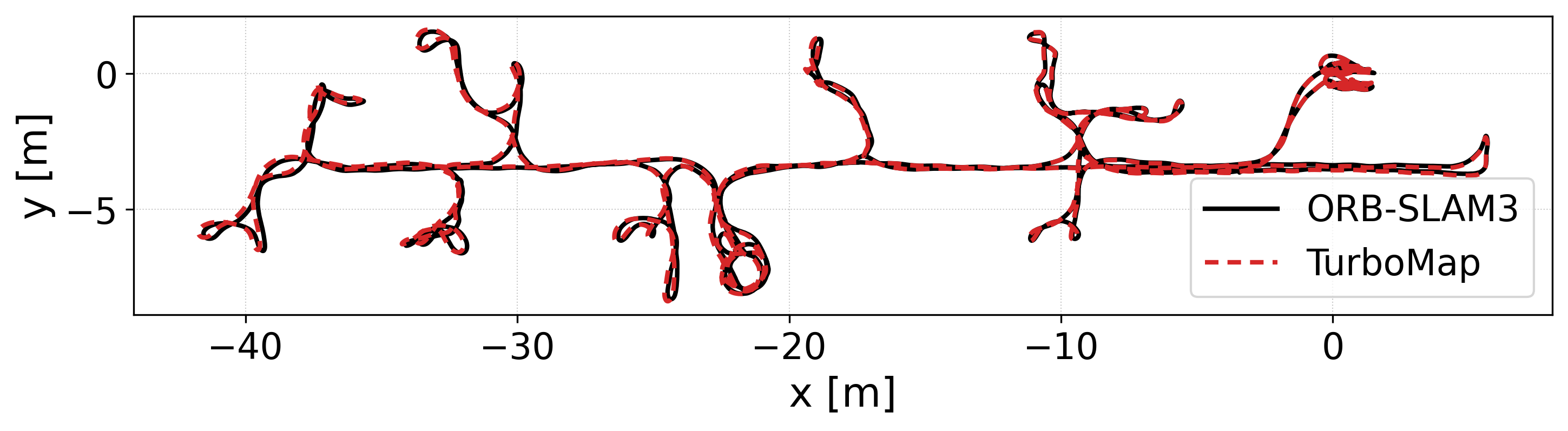}
    \caption{Trajectories: \SYS{} vs. ORB-SLAM3 (Corridor1)}
    \label{fig:trajectory_comparison}
    \vspace{-10pt}
\end{figure}

\subsection{Trajectory}
\label{eval:trajectory}
\autoref{tab:combined_results} compares the Absolute Trajectory Error (ATE) RMSE between \SYS{} and ORB-SLAM3. As shown, both systems achieve comparable accuracy across all sequences, demonstrating the robustness of \SYS{} in correctly processing keyframes. \autoref{fig:trajectory_comparison} visualizes the estimated trajectories for both systems in the corridor1 sequence. The trajectories closely align, indicating that \SYS{} preserves the localization accuracy of ORB-SLAM3.

\begin{table}[t!]
\renewcommand{\arraystretch}{1.4}
\vspace{10pt}
\centering
\setlength{\tabcolsep}{8pt}
\resizebox{1\linewidth}{!}{
\begin{tabular}{|c|c|c|c|c|}
\hline
\multirow{2}{*}{\textbf{Sequence}} & \multicolumn{2}{c|}{\textbf{TurboMap}} & \multicolumn{2}{c|}{\textbf{ORB-SLAM3}} \\
\cline{2-5}
 & \textbf{ATE} & \makecell{\textbf{\#Skipped LBA} \\ \textbf{\& KF Culling}} 
 & \textbf{ATE} & \makecell{\textbf{\#Skipped LBA} \\ \textbf{\& KF Culling}} \\
\hline
\textbf{MH01}        & {\small 0.040} & {\small 3.4} & {\small 1.105} & {\small 58} \\
\hline
\textbf{V101}        & {\small 0.039} & {\small 18.8} & {\small 0.091} & {\small 102.6} \\
\hline
\textbf{room1}       & {\small 0.011} & {\small 5.8} & {\small 1.005} & {\small 140.4} \\
\hline
\textbf{corridor1}   & {\small 0.045} & {\small 9.2} & {\small 10.136} & {\small 343.8} \\
\hline
\end{tabular}
}
\caption{Comparison of \SYS{} and ORB-SLAM3 with high keyframe insertion: ATE RMSE (m), skipped LBAs, and keyframe culling (avg. 5 runs, desktop setting).
% \kar{describe how many times you ran the experiment. Otherwise, it is confusing why the numbers are in fractions. }
}
\vspace{-10pt}
\label{tab:stress_test}
\end{table}

\begin{figure}[t]
    \centering
    \hspace*{-1cm}
    \includegraphics[width=0.65\linewidth]{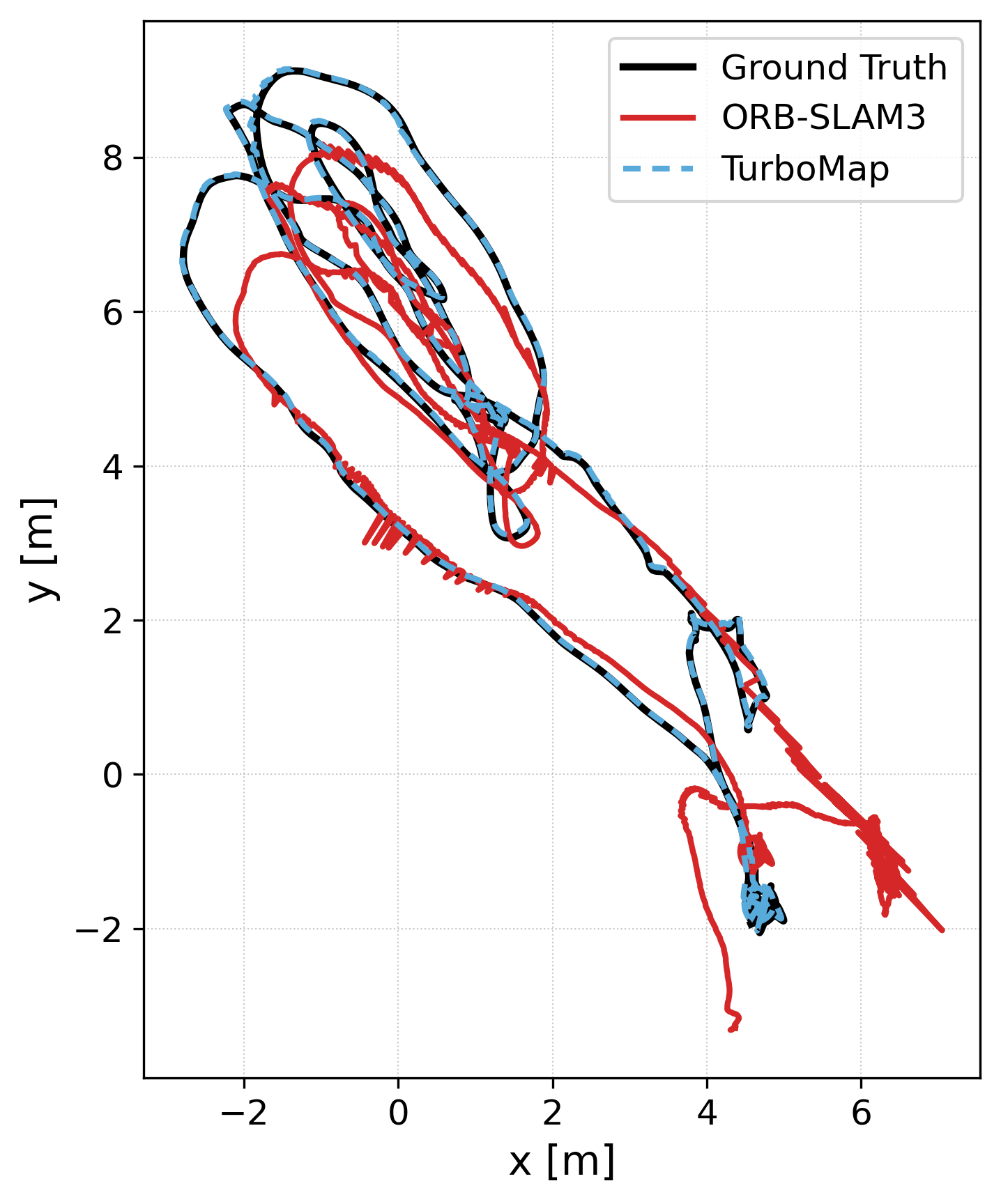}
    \caption{Trajectory comparison in MH01 between \SYS{} and ORB-SLAM3 under a high keyframe insertion rate.}
    \label{fig:MH01_high_load_trajectory}
    \vspace{-15pt}
\end{figure}

\subsection{Stress Testing Local Mapping}
\label{eval:stress_test}

To investigate system performance under rapidly changing scenes, we design an experiment that simulates scenarios requiring a high keyframe insertion rate to maintain tracking accuracy. This experiment enables us to evaluate how \SYS{} and ORB-SLAM3 perform under a heavy local mapping load. To do that, we simulate a sequence with high camera motion. Rapid camera motion can cause substantial changes in the observed scene between consecutive frames due to large viewpoint shifts and motion blur, resulting in previously tracked features becoming unobservable or unrecognizable. If keyframes are inserted too sparsely, the viewpoint difference between keyframes becomes large, making descriptor matching and triangulation less reliable. More frequent keyframe insertion shortens this difference, preserves sufficient feature overlap for robust matching, and helps maintain tracking accuracy under fast motion. Typically, more frequent keyframe insertion improves tracking robustness under challenging camera motion~\cite{Mur-Artal2015}.

To achieve higher keyframe insertion rates, we apply two modifications to both systems. First, ORB-SLAM3 includes a limiter that enforces real-time playback by pausing execution when tracking processes a frame faster than the sequence timestamp. We disable this limiter to accelerate frame processing and thereby simulate high camera velocity. Second, ORB-SLAM3 enforces a minimum frame interval between consecutive keyframe insertions. We reduce this threshold to permit denser keyframe insertion when rapid motion requires it. Together, these modifications increase the load on local mapping and enable a direct evaluation of the robustness and scalability of \SYS{} relative to ORB-SLAM3.

As shown in \autoref{tab:stress_test}, \SYS{} significantly outperforms ORB-SLAM3 under heavy load. While ORB-SLAM3 produces very high trajectory errors, \SYS{} maintains an accuracy comparable to normal load conditions, which is substantially lower than the error in ORB-SLAM3. This difference is also evident in \autoref{fig:MH01_high_load_trajectory}, where the trajectory produced by ORB-SLAM3 in the MH01 sequence diverges significantly from the ground truth, while \SYS{} produces a trajectory that closely aligns with it. This discrepancy can be explained by the way ORB-SLAM3 manages new keyframes. Local mapping maintains a fixed-size keyframe queue and processes keyframes sequentially. When the queue is not empty, ORB-SLAM3 skips LBA and Keyframe Culling stages for the current keyframe and starts processing the next keyframe in the queue in order to save time and avoid keyframe drops. However, skipping these stages impacts trajectory accuracy. \autoref{tab:stress_test} further illustrates this point by reporting the number of times LBA and Keyframe Culling are skipped under heavy load in both systems. The number of skips is substantially higher in ORB-SLAM3, which is the primary factor contributing to its decreased accuracy.

\section{Conclusion} \label{sec:conclusion}
In this work, we presented \SYS{}, a GPU-accelerated and CPU-optimized backend for ORB-SLAM3 that addresses the primary performance bottlenecks in the local mapping pipeline. Unlike prior work that mainly accelerates bundle adjustment, \SYS{} targets the remaining dominant components that are constrained by shared-state updates and synchronization. We restructure Map Point Creation to enable parallel Keypoint Correspondence Search on the GPU, redesign Map Point Fusion for efficient parallel execution, and optimize Keyframe Culling on the CPU. To support these GPU-parallelized stages efficiently, we introduce a persistent GPU-resident keyframe storage that minimizes data transfer and synchronization overhead. Together with the integration of an existing GPU-accelerated LBA solver, these changes enable efficient parallel execution while preserving correctness and system behavior. Experimental results show that \SYS{} achieves an average local mapping speedup of $1.3\times$ on the \EUROC{} dataset and $1.6\times$ on the \TUMVI{} dataset without degrading accuracy. These results demonstrate that holistically restructuring local mapping and carefully managing data movement can substantially improve the performance of modern visual SLAM systems.

\addtolength{\textheight}{-12cm}   % This command serves to balance the column lengths

%%%%%%%%%%%%%%%%%%%%%%%%%%%%%%%%%%%%%%%%%%%%%%%%%%%%%%%%%%%%%%%%%%%%%%%%%%%%%%%%

%%%%%%%%%%%%%%%%%%%%%%%%%%%%%%%%%%%%%%%%%%%%%%%%%%%%%%%%%%%%%%%%%%%%%%%%%%%%%%%%



% \section*{ACKNOWLEDGMENT}

% The preferred spelling of the word ÒacknowledgmentÓ in America is without an ÒeÓ after the ÒgÓ. Avoid the stilted expression, ÒOne of us (R. B. G.) thanks . . .Ó  Instead, try ÒR. B. G. thanksÓ. Put sponsor acknowledgments in the unnumbered footnote on the first page.

%%%%%%%%%%%%%%%%%%%%%%%%%%%%%%%%%%%%%%%%%%%%%%%%%%%%%%%%%%%%%%%%%%%%%%%%%%%%%%%%

% References are important to the reader; therefore, each citation must be complete and correct. If at all possible, references should be commonly available publications.

\bibliographystyle{IEEEtran}
\bibliography{IEEEabrv, references}

\end{document}